\documentclass[runningheads]{llncs}
\usepackage{graphicx}
\usepackage{comment}
\usepackage{amsmath,amssymb} 
\usepackage{color}
\usepackage{hyperref}  

\usepackage{enumitem}
\usepackage{gensymb}   
\usepackage{tabu,multirow}
\usepackage{booktabs}
\usepackage{bbding}
\usepackage{wrapfig}
\usepackage[skip=1pt]{caption} 
\newcommand{\CMB}{\CheckmarkBold}
\usepackage{caption}
\usepackage{amsmath}

\begin{document}
\pagestyle{headings}
\mainmatter
\def\ECCVSubNumber{6150}  

\title{Neural Mesh Refiner for 6-DoF Pose Estimation} 

\authorrunning{Di Wu et al.}


\author{Di Wu \and Yihao Chen \and Xianbiao Qi \and Yongjian Yu \and Weixuan Chen \and Rong Xiao}
\institute{PingAn Property \& Casualty Insurance Company}

\maketitle

\begin{abstract}

How can we effectively utilise the 2D monocular image information for recovering the 6D pose (6-DoF) of the visual objects?
Deep learning has shown to be effective for robust and real-time monocular pose estimation.
Oftentimes, the network learns to regress the 6-DoF pose using a naive loss function.
However, due to a lack of geometrical scene understanding from the directly regressed pose estimation, there are misalignments between the rendered mesh from the 3D object and the 2D instance segmentation result, \emph{e.g.}, bounding boxes and masks prediction.
This paper bridges the gap between 2D mask generation and 3D location prediction via a differentiable neural mesh renderer.
We utilise the overlay between the accurate mask prediction and less accurate mesh prediction to iteratively optimise  the direct regressed 6D pose information with a focus on  translation estimation.
By leveraging geometry, we demonstrate that our technique significantly improves direct regression performance on the difficult task of translation estimation and achieve the state of the art results on Peking University/Baidu - Autonomous Driving dataset and the ApolloScape 3D Car Instance dataset.
The code can be found at \url{https://bit.ly/2IRihfU}.
\end{abstract}

\section{Introduction}
The success of an autonomous agent is contingent on its ability to detect and localise the objects in its surrounding environment.
Understanding the 3D world from 2D images is one of the fundamental problems in computer vision.
Then the natural question is how much 3D spatial information one can infer from a single image?
Deep learning methods on monocular imagery have come a long way in recent years in understanding ill-posed inverse graphics problem like
6D pose tracking~\cite{li2018deepim,xiang2017posecnn,ma2019trafficpredict}, depth prediction~\cite{eigen2014depth,garg2016unsupervised,godard2017unsupervised,zhou2017unsupervised,rad2018feature}, or shape recovery~\cite{zhou2017unsupervised,kundu20183d}.


The seminal work of KITTI benchmark~\cite{geiger2012we} has led to the 3D bounding box detection from monocular imagery emerging as a much-researched topic and a menagerie of network architectures have been developed~\cite{chen2016monocular,mousavian20173d,kehl2017ssd,rad2017bb8,li2018deepim,simonelli2019disentangling}.
To date, the 3D object detection literature has been dominated by approaches which represent the 3D occupancy of an individual object with a 3D bounding box.
The research of 2D image detection has transitioned from a phase of a crude 2D bounding box detection~\cite{ren2015faster} to a finer mask prediction~\cite{he2017mask}.
In analogy to 2D object detection, we should strive to detect the 3D objects at a finer scale, such as voxels~\cite{choy20163d}, point clouds~\cite{fan2017point}, and meshes~\cite{wang2018pixel2mesh}.
To this end, we present a novel 3D instance detection algorithm which takes a single monocular RGB image as input and produce high-quality 3D polygon meshes for their compactness and geometric properties.

To reason the 3D instance from 2D images, the placement of 3D occupancy in the world coordinate should be in accordance with its rendering onto the 2D image plane.
This task can be dissected into inferring the 3D shape and locate the 3D rotation and translation information conforming to 2D cues.
For autonomous driving, the estimation of translation is of critical importance.
Particularly, the translation distance of traffic participants along longitudinal axis varies significantly.
Notwithstanding, Convolutional Neural Networks (CNNs) and the absolute position information are two concepts that rarely get discussed together.
A common consensus is that CNNs attain translational invariance (for the task of image classification), or translation equivariance (for the task of object detection or instance segmentation).
Hence, the accurate estimation of translation vector is an extremely challenging modality.

Current neural network architectures regress the translation vector by exploiting the position encoding in CNNs implicitly~\cite{zhou2019objects,kundu20183d} or explicitly~\cite{wu20196d}.
However, as observed in~\cite{song2019apollocar3d}, there are visible misalignments between the 2D  masks and 3D mesh rendering using the predicted rotation matrix $R$ and translation vector $T$.
This is because the direct regression approach doesn't take the geometric information when generating the 6D pose estimation.

In this work, we perform stage-wise training for accurate monocular 3D object detection.
The base model which encodes translational information explicitly in pixel space is trained for 2D instance detection and 6-DoF regression.
Then a geometric reasoning part is adopted to refine the 3D prediction so that the rendering  of 3D shape on the image plane is aligned with the more accurate 2D prediction.
Our contributions are summarised as follows:

\begin{itemize}[noitemsep,nolistsep]
\item We propose to optimise the shape information, rotation and translation regression via the modelling of homoscedastic uncertainty. This formulation enables to learn the required 3D information synergistically and optimally.
\item We leverage the more accurate 2D mask prediction to refine the regressed 3D prediction by a differentiable neural 3D mesh renderer.
      To facilitate the renderer to probe the plausible 3D state space, we propose a geometric state initialisation scheme.
\item We propose a novel model ensemble scheme that utilises IoU score between the 2D Mask and 3D Mesh projection in place of the traditional average predictions for the model ensemble.
\item We provide a critical review of the two metrics used to judge 3D object detection and advocate to merge the merits of the two metrics to be more adequately applicable under practical autonomous driving scenario.
\end{itemize}

\section{Related Work}


\noindent \textbf{Monocular-based 3D object detection}.
Deep3DBox~\cite{mousavian20173d} exploits the constraints from projective geometry to estimate full 3D pose and object dimensions from a 2D box. The central idea is that the perspective projection of a 3D bounding box should fit tightly to at least one side of the predicted 2D bounding box.
SSD-6D~\cite{kehl2017ssd} utilises the 2D detection hypothesis to estimate 6D pose with discretisations of the full rotational space. The training procedure is conducted on synthetically augmented datasets.
OFTNet~\cite{roddick2018orthographic} proposes an orthographic feature transform, mapping features extracted from 2D space to a 3D voxel map. By integrating along the vertical dimension, the voxel map features are reduced to 2D birds-eye view and the detection hypotheses are efficiently processed by exploiting integral-image representations.
Mono3D~\cite{chen2016monocular} focuses on the generation of 3D boxes, scored by different features like contour, shape, location priors and class semantics. A precondition for their test time inference is the requirement of semantic and instance segmentation results.
 proposes a novel loss formulation by lifting 2D detection, orientation, and scale estimation into 3D space.
ROI-10D~\cite{manhardt2019roi} uses a parallel stream based on the SuperDepth network~\cite{rad2018feature} which predicts per-pixel depth from the same monocular image.
The predicted depth maps are used to support distance reasoning in the 3D lifting part of their network.
The shape information is created by a projection of  truncated signed distances fields (TSDF)~\cite{curless1996volumetric}.

\noindent \textbf{Stage-wise training}.
FQNet~\cite{liu2019deep} infers the 3D IoU between the 3D proposals and the object solely based on 2D cues.
During the detection process, a large number of candidates in the 3D space are sampled and the best candidate is picked out by the spatial overlap between proposals and the object.
MonoGRNet~\cite{qin2019monogrnet} optimises the locations and poses of the 3D bounding boxes in the global context. The training is conducted stage-wise: first, the backbone is trained together with the 2D detector and next, geometric reasoning modules are trained.
The Deep iterative Matching method~\cite{li2018deepim} requires pre-training on a synthetic dataset using a disentangled representation of the 3D location and 3D orientation. The training process uses a FlowNet~\cite{dosovitskiy2015flownet} to predict a relative SE(3) transformation to match the observed and rendered image of an object.

\noindent \textbf{Position encoding in CNN}.
In contrast to fully connected networks, CNNs achieve efficiency by learning weights associated with local filters with a finite spatial extent.
An implication of this is that a filter may know what it is looking at, but not where it is positioned in the image.
Information concerning absolute position is inherently useful, and it is reasonable to assume that deep CNNs may implicitly learn to encode this information if there is a means to do so.
A coordinate transform problem is investigated in~\cite{liu2018intriguing}, showing the failure of CNNs in the task of coordinate mapping. Then a solution \emph{CoordConv} is proposed to allow networks to learn either complete translation invariance or varying degrees of translation dependence. Nevertheless, the paper also observes some degree of translation leakage from the network prediction.
Operating at a pixel level, ~\cite{de2017semantic} tackles the problem of semantic instance segmentation.
Similarly, based on a Fully Convolutional Network (FCN), ~\cite{wang2019solo} can learn decent  positional information without explicitly encoding absolute positional information.
For the task of depth estimation from single images,~\cite{van2019neural} found that networks ignore the apparent size of known obstacles in favour of their vertical position in the image.
A hypothesis~\cite{islam2020much} reveals a surprising degree of absolute position information that is encoded in commonly used neural networks: the padding near the border delivers position information to learn.

\noindent \textbf{3D Shape Renderer.}
For modelling the 3D world behind 2D images, a poly mesh is a promising candidate for its compactness and geometric properties.
However, it is not straightforward to model a polygon mesh from 2D images using neural networks because the conversion from a mesh to an image involves a discrete operation which prevents back-propagation.
Neural Mesh Render (NMR)~\cite{kato2018neural} is the first attempt to approximate the gradient for rasterisation that enables the integration of rendering into neural networks.
CMR~\cite{kanazawa2018learning} presents a learning framework for recovering the 3D shape, camera, and texture of an object from a single image based on the NMR.
Mesh R-CNN~\cite{gkioxari2019mesh} augments Mask R-CNN~\cite{he2017mask} with a mesh prediction branch that outputs meshes with varying topological structure.
C3DPO~\cite{novotny2019c3dpo} extracts 3D models of deformable objects from 2D keypoint annotations in unconstrained images.
~\cite{kulkarni2019canonical} explores the task of Canonical Surface Mapping (CSM). The key insight is that CSM task (pixel to 3D), when combined with 3D projection (3D to pixel), completes a cycle.
VoteNet~\cite{qi2019deep} leverages architectures in the 2D detector and proposes an end-to-end 3D object detection network based on a synergy of deep point set networks and Hough voting. 

\section{Method}

We address the problem of monocular 3D object detection, where the input is a single RGB image with a corpus of mesh models and the output consists  object 3D shape, 3-DoF rotation and 3-DoF translation, expressed in camera coordinates. Specifically, the output can be defined as a tuple $\boldsymbol{O}=(\mathcal{B}, \mathcal{M}, \mathcal{M}^m, R, T)$ with $\mathcal{B}$ as bounding box, $\mathcal{M}$ as mask from base model, $\mathcal{M}^m$ as mask produced by the mesh from rotation $R$ and translation $T$.
We first introduce the dual heads base model to predict 2D instance segmentation and 3D pose estimation with a novel weight learning scheme for balancing the losses from the multi-task.
Then, by investigating the misalignment between the predicted 3D mesh projected on camera plane and the 2D mask from the base model, we propose a neural mesh refiner to adjust 3D regression and a novel model ensemble scheme.

\subsection{Dual Heads Base Model}\label{sec:Dual Heads Base Model}

\begin{figure*}[t]
        \centering
        \includegraphics[width=1.0\textwidth]{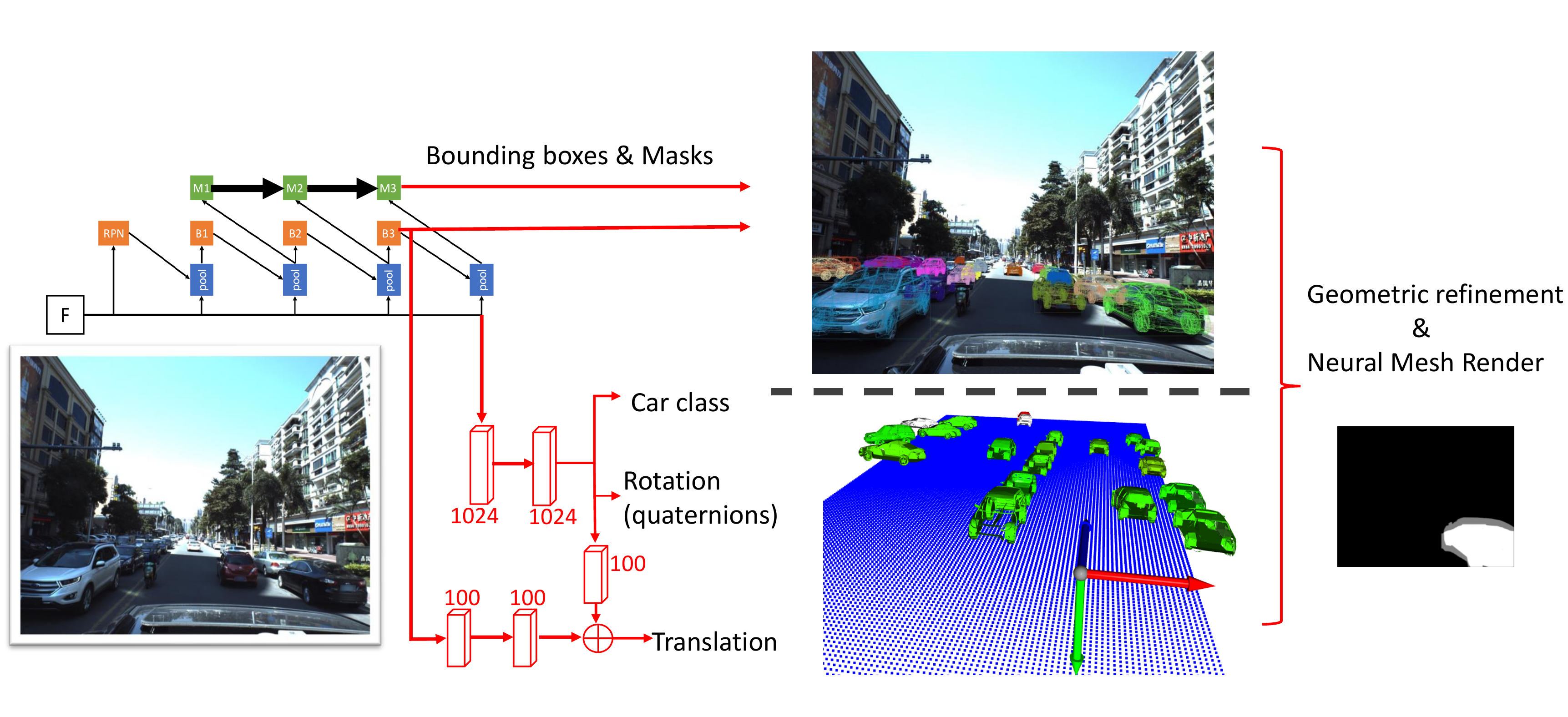}
        \caption{
        \textbf{Overview of the proposed framework.} A monocular image is taken as input to perform canonical instance segmentation (top) and estimate vehicles' 3D translation and rotation (bottom). $\oplus$ denotes the concatenation operation. The black branch represents a canonical multi-stage instance segmentation task and the red branch is the proposed branch for estimating object 6-DoF pose and its sub-category to provide 3D shape information.
        The refinement procedure takes the bounding box and mask as the more accurate signal and refined the translation accordingly.
        }
        \vspace{-0.5cm}
        \label{fig:system_pipeline}
\end{figure*}

The dual heads base model is built upon 6D-VNet~\cite{wu20196d} which is a task-specific canonical object detection network as shown in~\ref{fig:system_pipeline}.
The system is a multi-staged network trained end-to-end to estimate the 6-DoF pose information for the object of interest.
The first stage of the network is a multi-stage 2D object detection network.
The second stage of the network is the customised heads to estimate the object 6-DoF pose information and is split into two parts: the first part only takes RoIAlign from each candidate box if the candidate is vehicle and performs sub-class categorisation and rotation estimation.
Due to in-plane rotation is unique for a given vehicle class, all vehicles share similar rotational features for the same yaw, pitch, and roll angles.
Therefore, the fixed-size visual cue from RoIAlign layer is sufficient for estimating the candidate sub-category and rotation.
The second part takes both RoIAlign feature and bounding box information (in world unit as described in~\cite{wu20196d}) via a concatenation operation to estimate the 3-dimensional translational vector.
The joint feature combination scheme implicitly encodes the object class and rotation information via the concatenation operation.
The translation regression head functions in synergy when it is combined with the joint loss from sub-category classification and quaternion regression.
We minimise the following loss $\mathcal{L}$ to train our network in an end-to-end fashion:
\begin{equation}
\vspace{-0.1cm}
\mathcal{L}  = \mathcal{L}_{det} + \mathcal{L}_{inst}
\vspace{-0.1cm}
\end{equation}
where $\mathcal{L}_{det}$ denotes the multi-task loss as in~\cite{chen2019hybrid}: $\mathcal{L}_{det} = \sum^T_{t=1} (\mathcal{L}_{cls} + \mathcal{L}_{box} + \mathcal{L}_{mask})$ with number of stage $T=3$. The classification loss $\mathcal{L}_{cls}$, 2D bounding box loss $\mathcal{L}_{box}$ and 2D mask loss $\mathcal{L}_{mask}$ are identical as those defined in canonical detection network of Mask R-CNN~\cite{he2017mask}.

\begin{wrapfigure}{R}{0.5\textwidth}
  \begin{center}
    \vspace{-2cm}
    \includegraphics[width=0.48\textwidth]{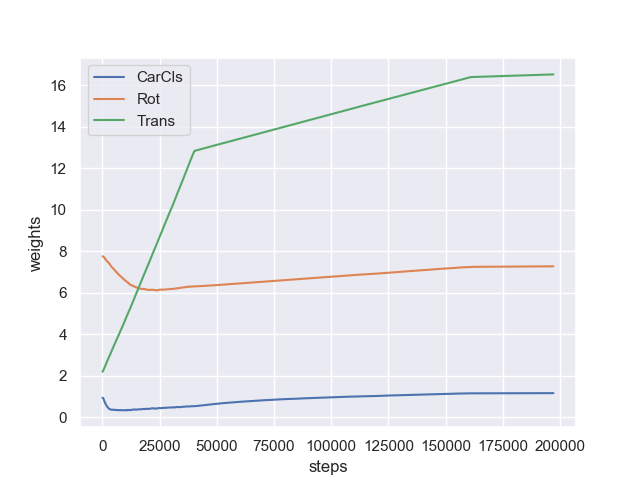}
        \caption{
       \textbf{Bayesian weight learning scheme.}
       The y-axis represents the weight ($\sigma^{-2}$) associates to the corresponding loss (we decrease the learning rate 10 folds at 40k and 160k steps).
       We initiate the weight for car classification, rotation, translation regress loss as $1, 8, 2$ so that the initial losses start within the same scale bracket. Since translation regression is the more difficult part of the regression task, the weight for translation loss steadily increases to 16.5, and the weight for rotation loss decreases to 7.25 and the weight for the cross entropy loss for car classification subdue stabilised around 1.2.
        }
        \label{fig:homoscedatic_uncertainty}
        \vspace{-1cm}
  \end{center}
\end{wrapfigure}

$\mathcal{L}_{inst}$ denotes the individual instance loss for 6-DoF estimation with losses from sub-class classification ($\mathcal{L}_{shape}$), rotation regression ($\mathcal{L}_{rot}$), and translation regression $(\mathcal{L}_{trans})$ as defined in~\cite{wu20196d}.
Previous works used scaling factors to balance the losses. This hyper-parameter attempts to keep the expected value of errors approximately equal.
However, the choosing such hyper-parameter is \emph{ad-hoc} and requires significant tuning to get reasonable results.
We adopt the Bayesian learning scheme~\cite{kendall2017geometric} for optimal weight learning to simultaneously regress position and orientation and classify sub-class shape as:
\begin{equation}
\mathcal{L}_{inst}  =  \mathcal{L}_{shape} \cdot \sigma^{-2}_{shape} + \log \sigma^{2}_{shape} +  \mathcal{L}_{rot}\cdot \sigma^{-2}_{rot} +   \log \sigma^{2}_{rot} + \mathcal{L}_{trans}\cdot \sigma^{-2}_{trans} +  \log \sigma^{2}_{trans}
\end{equation}\label{eq:inst_loss}

The individual loss consists of two components: the residual regression and the uncertainty regularisation terms. The variance $\sigma^2$ is implicitly learnt from the loss function.
As the variance is larger, it has a tempering effect on the residual regression term; larger variances (uncertainty) results in a smaller residual loss. The second regularisation term prevents the network from predicting infinite uncertainty (and therefore zero loss).
We optimise the homoscedastic uncertainties, $\sigma^{-2}_{shape}, \sigma^{-2}_{rot}, \sigma^{-2}_{trans}$ through backpropagation with respect to the loss functions.
Consequently, we do not force to optimise all terms equally at the same time, but let the network decide its focus during training.
Fig.~\ref{fig:homoscedatic_uncertainty} shows the learning process of the loss weighting factors ($\sigma^{-2}$).
As it is expected, the task of translation regression is the most difficult task with a larger noise in output space. The weighting factor gradually learns to balance the loss without the necessity of sub-optimal manual tuning.

\subsection{Neural 6-DoF Mesh Refiner}\label{NMR}

We now illustrate how to adjust the directly regressed 3D translation and rotation according to 2D cues from the mask prediction via a novel neural mesh refiner.

In a paraxial refraction camera model, a point  $P=(x, y, z)$ in 3D world space can be projected to 2D point $P^{\prime}=(x^{\prime}, y^{\prime})$ in the image plane.
The relationship between a point in 3D world space $P=(x, y, z)$ and its image coordinates $P^{\prime}=(x^{\prime}, y^{\prime})$ by a matrix vector can be represented in homogeneous coordinate system as $P=(x, y, z, 1)$ and  $P^{\prime}=(x^{\prime}, y^{\prime}, 1)$.
Given camera intrinsic calibration matrix $K = [f_x ,0,c_x ;0,f_y ,c_y ;0,0,1]$ where $(f_x, f_y)$ are focal length of camera and $(c_x, c_y)$ are coordinates of the optical center of the camera, and rotation matrix $R$ and translation vector $T$ as extrinsic parameters.
Thus, the mapping of the 2D point to 3D world space can be formulated as follows:
 \vspace{-0.2cm}
\begin{equation}\label{eq:camera_intrinsics}
\left[\begin{array}{l}{x^{\prime}} \\ {y^{\prime}} \\ {1}\end{array}\right]=\left[\begin{array}{cccc}{f_{x}} & {0} & {c_{x}} & {0} \\ {0} & {f_{y}} & {c_{y}} & {0} \\ {0} & {0} & {1} & {0}\end{array}\right]\left[\begin{array}{cc}{R} & {T} \\ {\overrightarrow{0}} & {1}\end{array}\right]\left[\begin{array}{cc}{x} \\ {y} \\ {z} \\ {1}\end{array}\right]
\end{equation}\label{eq:projection_matrix}
 \vspace{-0.2cm}
A 3D mesh consists of a set of vertices $\left\{\boldsymbol{v}_{1}^{o}, \boldsymbol{v}_{2}^{o}, \ldots, \boldsymbol{v}_{N_{v}}^{o}\right\}$ where $N_v$ denotes the number of vertices and faces  $\boldsymbol{v}_{i}^{o} \in \mathbb{R}^{3}$ $\left\{\boldsymbol{f}_{1},\boldsymbol{f}_{2}, \ldots, \boldsymbol{f}_{N_{f}}\right\}$ where $N_f$ denotes number of faces.
$\boldsymbol{v}_{i}^{o} \in \mathbb{R}^{3}$represents the position of the $i$-th vertex in the 3D object space and $\boldsymbol{f}_{j} \in \mathbb{N}^{3}$ represents the indices of the three vertices corresponding to the $j$-th triangle faces. To render this object, vertices $\left\{\boldsymbol{v}_{i}^{o}\right\}$ in the object space are transformed into vertices $\left\{\boldsymbol{v}_{i}^{s}\right\}, \boldsymbol{v}_{i}^{s} \in \mathbb{R}^{2}$ in the screen space. This transformation is represented by the combination of differentiable transformations as defined in Eq.~\ref{eq:projection_matrix}.

\begin{figure}[t]
        \centering
        \includegraphics[width=0.9\textwidth]{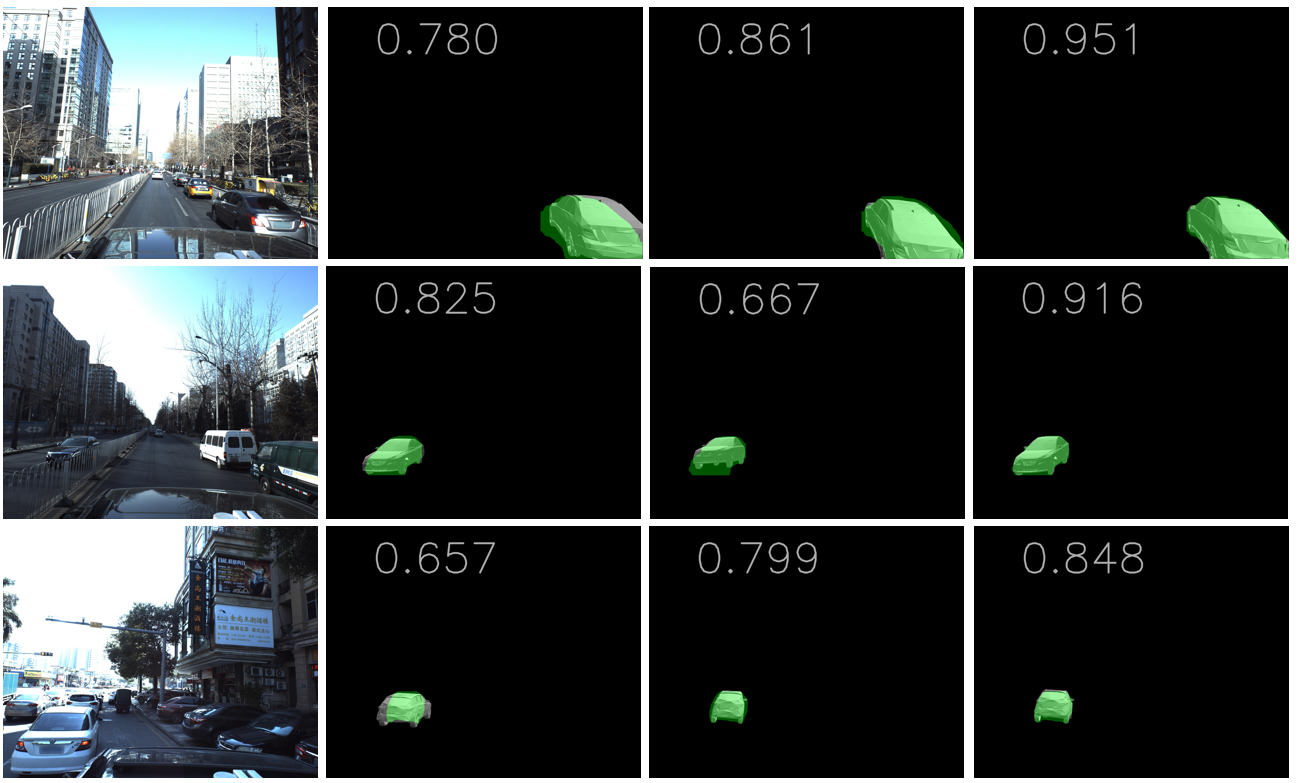}
        \caption{
        \textbf{Visualisation for Neural 6-DoF Mesh Refiner optimisation.}
        The learning process takes $R$ and $T$ as the optimising variables. Left-most is the original RGB image. The green mask represents the mask prediction from the base model and is served as the weakly supervised signal. The rendered reference image is optimised to match the green mask.  The numbers denote IoU score $\mathcal{S}^{mm}$ between mask $\mathcal{M}^m$  from 3D mesh and mask $\mathcal{M}$ from the base model, reflecting the accuracy of our 6-DoF pose estimation. Note that in the third row, since the ground truth mask is generated from the mesh projected onto 2D, the mask generated from the base model is also complete albeit the car is occluded.
        }
        \vspace{-0.5cm}
        \label{fig:NMR}
\end{figure}
\noindent \textbf{Instance Losses}. The goal is to match 3D mesh generated from $R$ and $T$ with 2D mask predictions $\mathcal{M}$.
The mask loss is defined as:
\begin{equation}\label{eq:mask_loss}
L_{\mathrm{mask}}=\sum_{i}\left\|\mathcal{M}-\mathcal{R}\left(\boldsymbol{v}_{i}, \boldsymbol{f} | R, T\right)\right\|_{2}
\end{equation}
The rasterisation process $\mathcal{R}(\cdot)$  of a single triangle is a discrete operation. If the centre of a pixel $P^{\prime}_j$ is inside of the face, the colour $I_j$ of the pixel $P^{\prime}_j$ becomes the colour of the overlapping face $I_{i,j}$.
Given supervised signal is the mask $\mathcal{M}$, then $I_{i,j}$ is independent of $\boldsymbol{v}_i$, $\frac{\partial I_{j}}{\partial \boldsymbol{v}_{i}}$ is zero almost everywhere. This means that the error signal back-propagated from a loss function to pixel $P^{\prime}_j$ does not flow into the vertex $\boldsymbol{v}_i$.
We adopt the approximate gradient for rasterisation~\cite{kato2018neural} that enables the integration of rendering into neural networks.

When training the refiner, we can either allow both rotation and translation to be incorporated into the optimisation process, or fix rotation and refine translation or vice versa.
Typically a Lie algebra is adopted for rotation representation~\cite{do2018real}.
Because vehicles are rigid objects, the initial rotation estimation of quaternion regressed from the RoIAlign feature is of high accuracy comparing with the regressed translation.
Under the scenario of autonomous driving where vehicles on the road lie on the same flat surface, translation distance of traffic participants along longitudinal axis varies significantly.
Therefore, freely allowing rotation in the optimisation process to alter the iteratively updated mesh will result in overfitting of the network and hurt the accurate 3D localisation.
Consider that for the distant objects, allowing the network to rotate in order to match the weak supervised signal of mask instead of optimising the more difficult task of translation refinement could generate unrealistic 3D pose.
Hence, during the refinement process, we fix the rotation matrix with an initial state of $R_0$ and only allow translation vector to be in the process of optimisation. We also observe that the IoU loss is more stable and converges faster than the grayscale loss. Therefore, we use the $L_{\mathrm{mask}}=  - IoU (\mathcal{M}, \mathcal{R}(\boldsymbol{v}, \boldsymbol{f}, R_0 |T))$ for stabilising training and faster convergence.
Fig~\ref{fig:NMR} illustrates the optimisation process.

\begin{wrapfigure}{R}{0.50\textwidth}
  \begin{center}
          \vspace{-2cm}
    \includegraphics[width=0.48\textwidth]{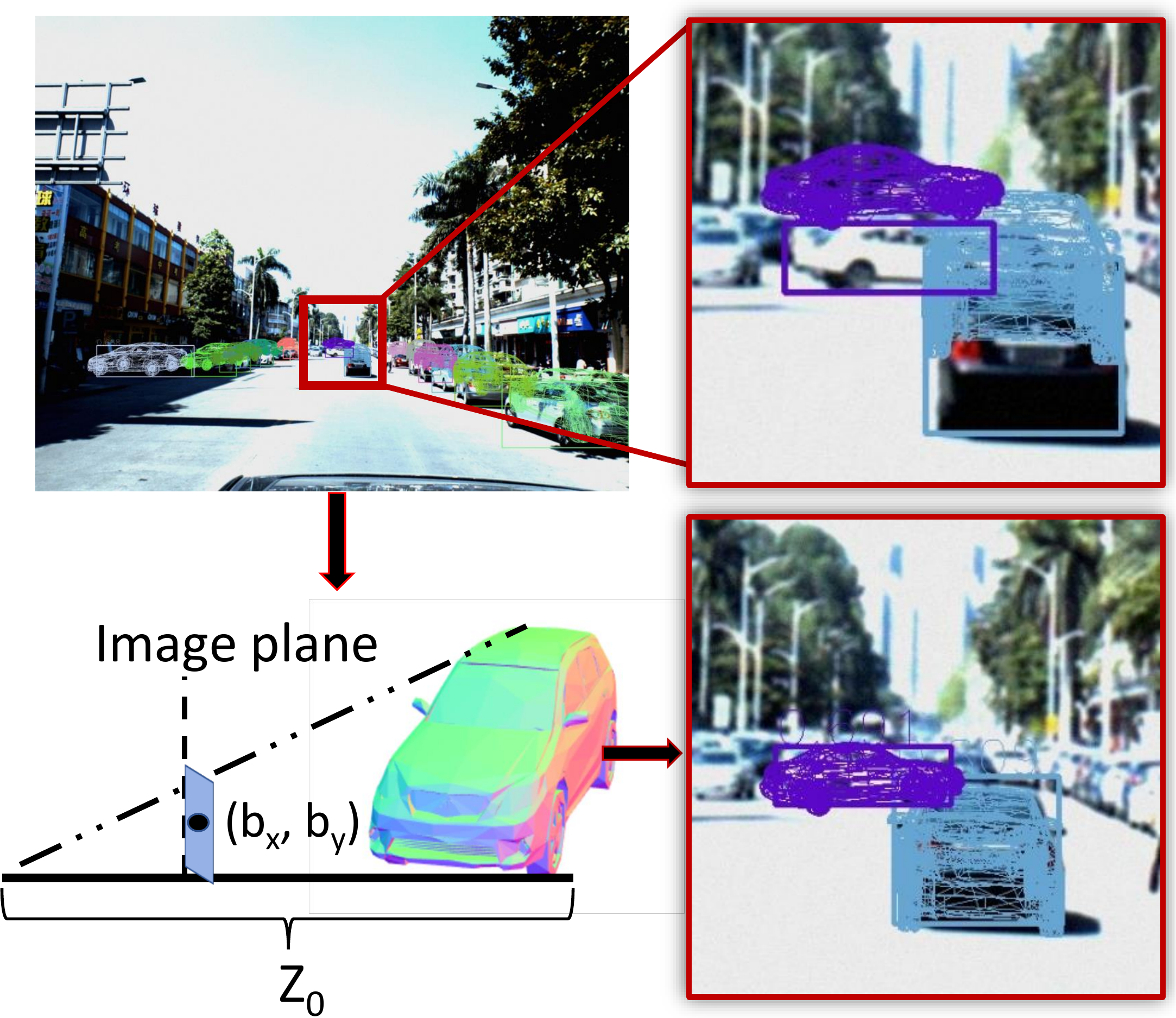}
       \caption{
        \textbf{Geometric restoration of the initial state.}
        The refinement procedure takes the bounding box and mask as the more accurate signals and modifies the translation according to Eq.~\ref{eq:alternative_initial_state}.
        Top: base model regression result; bottom: geometric restoration result.
        }
        \vspace{-1cm}
        \label{fig:geometric_state_initialisation}
  \end{center}
\end{wrapfigure}
\subsection{Geometric State Initialisation}\label{GSI}


One issue of the neural mesh refiner is that the initial state of translation $T_0=(x_0, y_0, z_0)$ should reside in a space that the network can probe a better state with a given step size.
Otherwise, if the initial state with $T_0$ generates a mesh that does not overlap with the 2D mask, the network will not be able to optimise with the given probing budget.
 At inference time, our dual heads based model (Sec.~\ref{sec:Dual Heads Base Model}) generates reliable bounding boxes in 2D pixel coordinates.
 In order to provide a valid initial state where the mesh overlaps with mask, we deploy the following trick to obtain an alternative initial state $\hat{T_0}=(\hat{x_0}, \hat{y_0}, z_0)$. Assuming the initial estimation $z_0$ is close enough to the truth state, we modify the $\hat{x_0}$ and $\hat{y_0}$ so as to match the predicted bounding box as (\emph{c.f.} Fig.~\ref{fig:geometric_state_initialisation}):
 \begin{equation}\label{eq:restore_x_y}
\hat{x_0}=\frac{z_0(b_x-c_x)}{f_x},\quad \hat{y_0}=\frac{z_0(b_y-c_y)}{f_y}
 \end{equation}\label{eq:alternative_initial_state}
 where $b_x,b_y$ denote centre coordinates of the detected bounding box in pixel space and $(f_x, f_y), (c_x, c_y)$ are the aforementioned focal length of camera and optical center of the camera.

 To decide whether the alternative initial state $\hat{T_0}$ from Eq.~\ref{eq:alternative_initial_state} is a better candidate than $T_0$, we use the IoU between the mask generated from mesh and mask generate from the base model as the criterion.
 Specifically: the 3D car model with predicted 6-DoF information can be projected onto the image coordinates and produce mesh mask $\mathcal{M}^m$. The base model mask branch predicts the mask segmentation which we denote as $\mathcal{M}$. The IoU score $\mathcal{S}^{mm}$ between $\mathcal{M}^m$ and $\mathcal{M}$ reflects the accuracy of our 6-DoF pose estimation. The higher IoU score, the more accurate is the estimated 6-DoF pose. If the IoU improves, then it's a better initial state and will be adopted for the mesh refiner.

\vspace{-0.3cm}
\subsection{Geometric Weighted Model Ensemble}\label{GWME}

\begin{figure*}[t]
        \centering
        \includegraphics[width=0.7\textwidth]{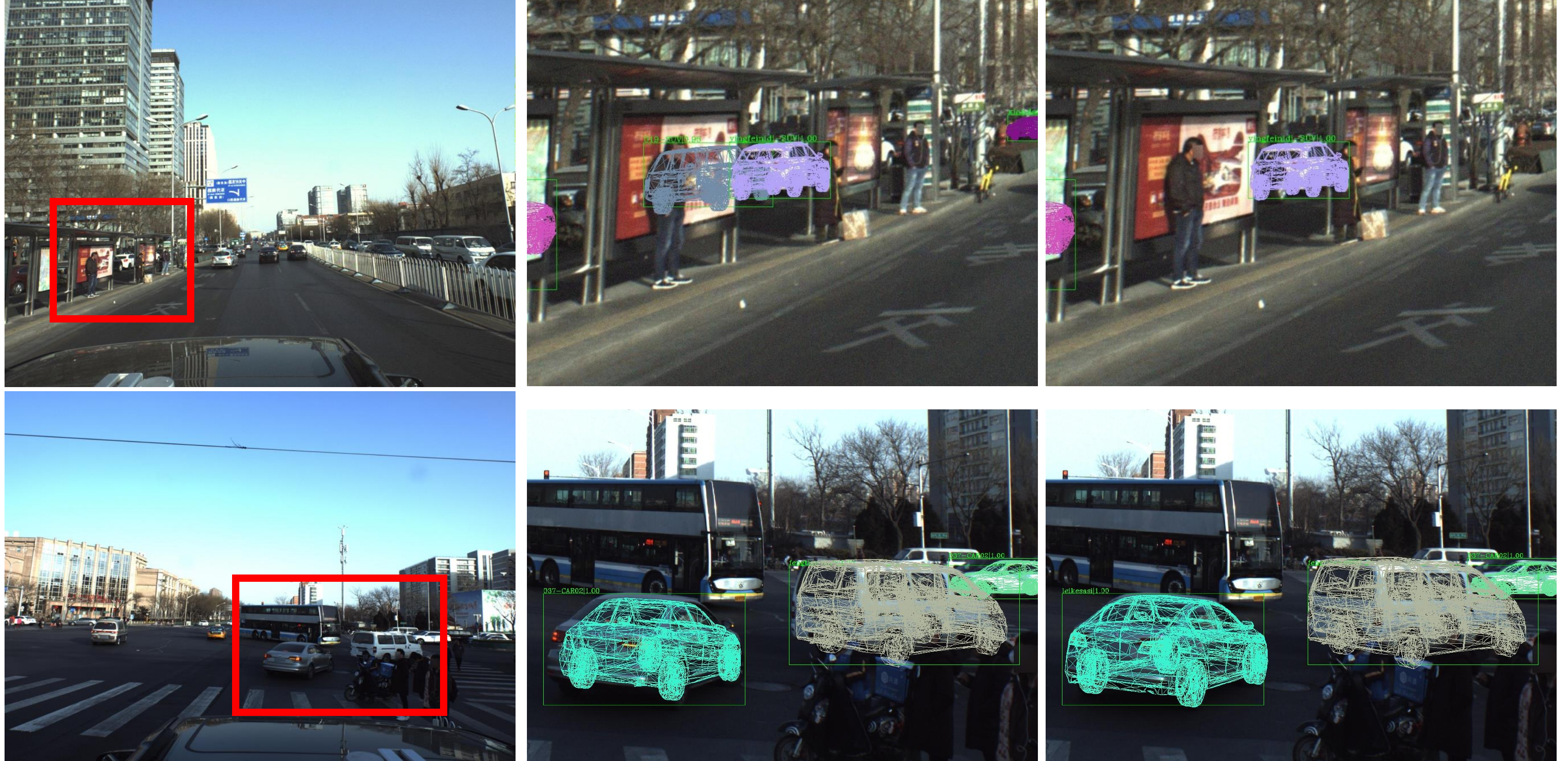}
        \caption{
       \textbf{ \emph{Top}: effect of model voting.} Middle represents model ensemble predictions without model voting scheme. Right is the output of our model voting scheme to suppress the spurious false detection.
        \textbf{\emph{Bottom}: effect of model ensemble schemes.} Middle represents average prediction and right is the geometric weighted model ensemble. The geometric weighted scheme takes the IoU score $\mathcal{S}^{mm}$  into consideration and the ensemble scheme is visually more in line with the overall mask prediction.
        }
         \vspace{-0.8cm}
        \label{fig:model_ensemble}
\end{figure*}

Model ensemble is a well unknown technique for reducing the variances of different models. Traditionally, for model ensemble, mean average prediction is adopted.
Due to the multi-task nature of the proposed module, we propose a novel way of model ensemble.
Suppose we have outputs $\left\{\boldsymbol{O}_{1}, \boldsymbol{O}_{2}, \ldots, \boldsymbol{O}_{N_{m}}\right\}$ where $N_m$ denotes the number of models (\emph{e.g.}, training with different seeds, input sizes, regression losses, \emph{etc.}).
We firstly select the highly confident predictions from all models based on bounding box non-maxima suppression.
In order not to include excessive false positive predictions from model outliers, we adopt a \emph{model voting} scheme: a vote number count $N_v$ (we set $N_v=N_m$ in our experiments) to retain the confident predictions whose overlapping bounding boxes count from distinctive models is no smaller than $N_v$.
 The false positives are suppressed as shown in Fig.~\ref{fig:model_ensemble}(top).

The merged translation prediction $\bar{T}$ can be the mean average of $T_i$ from model $i$.
Nonetheless, the IoU score $\mathcal{S}^{mm}$ between $\mathcal{M}^m$ and $\mathcal{M}$ reflects the accuracy of our 6-DoF pose estimation.
Hence, we propose the following scheme to utilise the correspondence between 2D and 3D with visually more aligned mesh generated from $T, R$:
\begin{equation*}
 \bar{T} = \frac{\sum^{N_m}_i (\mathcal{S}^{mm}_i \cdot T_i)}{\sum^{N_m}_i \mathcal{S}^{mm}_i}
\end{equation*}
%
%
%

\section{Experiments}

\begin{wrapfigure}{R}{0.4\textwidth}
  \begin{center}
    \vspace{-2cm}
    \includegraphics[width=0.38\textwidth]{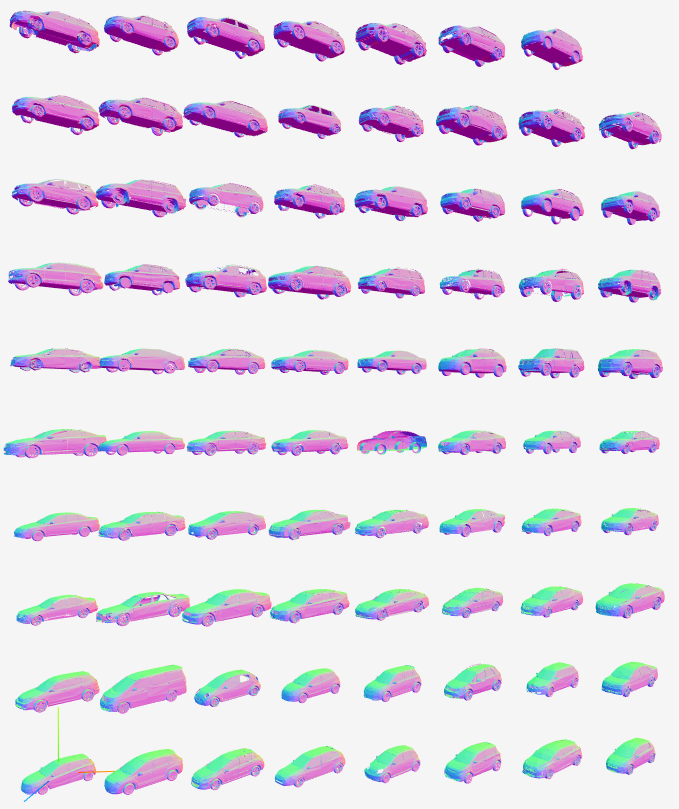}
        \caption{79 car model meshes.}
        \label{fig:car_mesh}
        \vspace{-1cm}
  \end{center}
\end{wrapfigure}

In this section, we describe our evaluation protocol, compare to the state of the art for RGB-based approaches, and provide an ablative analysis discussing the merits of our individual contributions.
We also review the two metrics that are adopted to rank the performance and present our critical review for the future practical deployment evaluation.

\noindent \textbf{Implementation details:} \label{sec:implementation_details}
We build our framework based upon the open source project MMDetection~\cite{mmdetection}. Pixel-level transform for image augmentation is called from the Albumentations~\cite{2018arXiv180906839B} library.
The detector is a 3-stage Hybrid Task Cascade (HTC)~\cite{chen2019hybrid} and the backbone is ImageNet pretrained High-resolution networks(HRNets)~\cite{SunXLW19}.
We use the Adam optimiser with a starting learning rate of $3e^{-3}$ with linear warm-up . We train the model for 120 epochs and anneal the learning rate at 80 and 100 epoch by 0.1. Since the monocular RGB images are of pixel size $2710 \times 3384$ and the vehicles only appear at the bottom part of the image, we crop out the top part of the image from 0 to 1480 pixels.

For Neural 6-DoF Mesh Refiner, we train for 20 epoches with a learning rate of 0.05 to refine the 3D translational vector.
The optimisation process will terminate if the IoU score $\mathcal{S}^{mm}$ exceeds 0.95.
Otherwise, we keep the optimised result corresponding to the highest $\mathcal{S}^{mm}$  at the end of the optimisation process.

\subsection{Datasets: Apolloscape Car Instance \& PKU/Baidu}
\emph{\textbf{Apolloscape 3D Car Instance dataset}}~\cite{kaggle_pku_baidu}\footnote{\url{http://apolloscape.auto/car_instance.html}} contains a diverse set of video sequences recorded in the street scenes from different cities.
Car models are provided in the form of triangle meshes. The mesh models have around $N_v=4000$ vertices and vertices and $N_f=5000$  triangle faces.
There are total 79 car models in three categories (sedan1, sedan2, SUV) with only 34 car models appearing in the training set (\emph{c.f.} Fig~\ref{fig:car_mesh}).
In addition, ignored marks are provided as unlabelled regions and we only use the ignored masks to filter out detected regions during test.
There are 3941/208/1041 high quality annotated images in the training-validation-test set.

\noindent \emph{\textbf{Peking University/Baidu - Autonomous Driving dataset}}~\cite{ma2019trafficpredict}\footnote{\url{https://www.kaggle.com/c/pku-autonomous-driving}}
is an extension of the Apolloscape 3D Car Instance dataset with an emphasis on translation prediction in a relative scale.
There are in total 4262 images for training and 2021 images for testing.

\subsection{Evaluation Metrics: \emph{A3DP-Abs}  \& \emph{A3DP-Rel}} \label{sec:evaluation_Apolloscape}
The evaluation metrics follow similar instance mean AP as the MS-COCO~\cite{lin2014microsoft}.
However, due to 3D nature, the 3D car instance evaluation has its own idiosyncrasies:
instead of using 2D mask IoU to judge a true positive, the 3D metric used in this dataset contains the perspective of shape $(s)$, 3D translation $(t)$ and 3D rotation $(r)$.
The shape similarity score is provided by an $N_{c} \times N_{c}$ matrix where $N_{c}$ denotes the number of car models.
For 3D translation and 3D rotation, the Euclidean distance and $arccos$ distance are used for measuring the position and orientation difference respectively.

Specifically, given an estimated 3D car model in an image $C_i=\{s_i, t_i, r_i\}$ and ground truth model $C_i^*=\{s_i^*, t_i^*, r_i^*\}$, the evaluation for these three estimates are as follows:
for 3D shape, reprojection similarity is considered by putting the model at a fix location and rendering 10 views $(v)$ by rotating the object. Mean IoU is computed between the two poses $(P)$ rendered from each view. Formally, the metric is defined as: $c_{shape} = \frac{1}{|V|} \sum_{v \in V} IoU (P(s_i), P(s^*_i))_v$, where $V$ is a set of camera views.
For 3D translation and rotation, the evaluation metric follows that of the canonical self-localisation: $c_{trans}=\parallel t_i - t^*_{i}\parallel^2$ and $c_{rot}=\arccos (|q(r_i)\centerdot q(r^*_i)|)$.

Then, a set of 10 thresholds from loose criterion to strict criterion ($c_0, c_1, \ldots, c_9$) is defined as:
$
    shapeThrs  - [.5:.05:.95],
    rotThrs    - [50:  5:  5],
    transThrs  - [2.8:.3:0.1]
$
where the most loose metric $c_0$: $0.5, 50, 2.8$ means shape similarity $>0.5$, rotation distance $< 50\degree$ and translation distance $<2.8$ metres, and stricter metrics can be interpreted correspondingly: all three criterion must be satisfied simultaneously so as to be counted as a true positive.

The precise translational estimation requirement is the major factor for the network to produce incorrect false positives, which is a challenging task from a human perspective as well.
The absolute translation thresholds are denoted as \emph{``A3DP-Abs" }.
In the Kaggle PKU/Baidu dataset, rather than evaluating the distance in the absolute distance, the ``AbsRel" commonly used in depth evaluation~\cite{geiger2012we} is adopted.
Formally, the criteria of  $c_{trans}$ is changed to $c_{trans}=\parallel t_i - t^*_{i}\parallel^2 / t^*_i$ and the threshold is set to $transThrs = [0.10:0.01:0.01]$.
The metrics with relative translation thresholds denotes as \emph{``A3DP-Rel"}.

\subsection{State-of-the-art comparison}
\noindent \textbf{Apolloscape Car Instance dataset:}
We compare with the state-of-the-art detectors of \emph{A3DP-Abs}  criterion in Tab.~\ref{tab:res_apolloscape}.
The baseline method 6D-VNet~\cite{wu20196d} extends Mask R-CNN by adding customised heads for predicting vehicle's finer class, rotation and translation with non-local module as an attention mechanism.
 3D-RCNN~\cite{kundu20183d} provides regression towards translation, allocentric rotation, and car shape parameters.
The baseline is further extended by adding mask pooling (MP) and offset flow (OF)~\cite{song2019apollocar3d}.  They together help voiding geometric distortion from regular RoI pooling and bring attention mechanism to focus on relevant regions.
The \emph{``human"} performance is obtained by the asking the human labeller to re-label the key points, and then passing them through the context-aware 3D solver.
Comparing with the direct regression approach, our approach achieves $1.4\%$ absolute mean AP improvement.

%

\begin{table*}
 \vspace{-0.5cm}
   \centering
   \footnotesize{
        \begin{tabu}{@{}lcccccc@{}}               \toprule
        [-1pt] \tabucline[1pt]{1-7}
            Method                              & Mask      & wKP     & Time(s)   & \multicolumn{3}{c}{A3DP-Abs}            \\
                                                                [-1pt] \tabucline[0.5pt]{5-7}
                                                &           &        &          & mean   &  c-0    & c-5                  \\
            \hline
            6D-VNet~\cite{wu20196d}             & pred      & -      & 0.42s    & 14.83  & 35.30   & 20.90           \\
            3D-RCNN~\cite{kundu20183d}          & gt        & -      & 0.29s    & 16.44  & 29.70   & 19.80             \\
            3D-RCNN + MP + OF~\cite{song2019apollocar3d}& pred      & -      & 0.34s    & 15.15  & 28.71   & 17.82             \\
           \hline
            \textbf{Ours}                       &  pred      &   -    & 0.33s    & 16.23  & 35.44  &22.68   \\
            \textbf{Ours} (Model Ensemble)      &  pred      &   -    & -        &\textbf{ 17.33}  &\textbf{ 36.16}  &\textbf{ 23.79 }    \\

            \hline
            Human                               & gt         & \CMB   & 607.41s  & 38.22 &  56.44  & 49.50      \\
        [-1pt] \tabucline[1pt]{1-7}
        \end{tabu}
}
    \caption{
    {\footnotesize Comparison among baseline algorithms. ``Mask" means the provided mask for 3D understanding (``gt” means ground truth mask and ``pred” means Mask-RCNN mask).
    ``wKP” means using keypoint predictions.
    ``MP” stands for mask pooling and ``OF” stands for offset flow.
    ``Times(s)” indicates the average inference times cost for processing each image.
    $c_0$ is the most loose criterion for evaluating AP and $c_5$ is in the middle of criterion.
    Note, the implementation of 3D-RCNN adopts the ground truth mask prediction during inference.
    }
} \label{tab:res_apolloscape}
 \vspace{-1cm}
\end{table*}

\noindent \textbf{PKU/Baidu dataset:}
We compare with the public solutions for evaluating  \emph{A3DP-Abs}  criterion in Tab.~\ref{tab:res_kaggle}.
QNet~\cite{qNET} is a two-stage detector that firstly detects 2D bounding boxes using Faster R-CNN~\cite{ren2015faster} and then a separate network for 6D-pose regression.
CenterNet~\cite{zhou2019objects} is an anchor-free object detector that uses keypoint estimation to find center points and regresses all other object properties.
CenterNet along has trouble regressing translation vector. CoordConv~\cite{liu2018intriguing} is a solution that give convolution access to its own input coordinates through the use of extra coordinate channels.


\begin{table*}
 \vspace{-0.5cm}
   \centering
   \small{
        \begin{tabu}{@{}lcc@{}}               \toprule
        [-1pt] \tabucline[1pt]{1-3}
            Method                                  &  \multicolumn{2}{c}{mean average precision}                     \\
                                                    [-1pt] \tabucline[0.5pt]{2-3}
                                                    & public LB   &  private LB                 \\
           \hline

           6D-VNet~\cite{wu20196d}           &    0.110     & 0.105   \\
           CenterNet~\cite{zhou2019objects}         &    0.118     & 0.112        \\
           QNet~\cite{qNET}                        &  0.124        & 0.116  \\
           CenterNet~\cite{zhou2019objects} + CoordConv~\cite{liu2018intriguing}     &  0.136       &  0.133      \\
           \textbf{Ours} (single model)                &  \textbf{ 0.142}    & \textbf{0.135 } \\
        [-1pt] \tabucline[1pt]{1-7}
        \end{tabu}
}
    \caption{
    {\footnotesize Results on Kaggle PKU/Baidu Dataset.}
     \vspace{-1cm}
} \label{tab:res_kaggle}
\end{table*}

\subsection{Ablation studies}
We present the ablation studies of our proposed module in the progressive tabular forms of Tab.~\ref{tab:ablation_apolloscape} and Tab.~\ref{tab:ablation_kaggle}.
We denote the abbreviations of different module as follows:
\emph{DHBM} as the dual-head base model in Sec.\ref{sec:Dual Heads Base Model} which is a improvement of 6D-VNet~\cite{wu20196d} with stronger backbone structure and Bayesian weight learning;
\emph{GSI} as Geometric State Initialisation in Sec.~\ref{GSI};
\emph{NMR} as Neural 6-DoF Mesh Refiner in Sec.~\ref{NMR};
\emph{GWME} as Geometric Weighted Model Ensemble in Sec.\ref{GWME}.

\noindent \textbf{Bayesian weight learning:} Row 1 \& 2 in Tab.~\ref{tab:ablation_kaggle} demonstrate that the coefficient learning paradigm is effective in updating the weights for multi-task losses.

\noindent \textbf{Neural 6-DoF Mesh Refiner:}Row 4 \& 5 in Tab.~\ref{tab:ablation_kaggle} show that by including  $R$ into optimisation process will lead to overfitting of the neural mesh refiner network. For example, the refiner will allow car flip so as to match the weakly supervised signal of silhouette. Fixing $R$ will direct the refiner to focus on updating translational vector.


\begin{table*}[t]
\scriptsize{
    \centering%
    \begin{tabular}{cccc|cccccccccccc}
     \toprule
   DHBM    & GSI    & NMR    & GWME  & $mAP$ & $c_0$ & $c_5$ & $AP^S$& $AP^M$& $AP^L$& $AR^1$& $AR^{10}$ & $AR^{100}$ & $AR^S$& $AR^M$ & $AR^L$ \\ \hline
      \CMB &       &         &          &0.162  & 0.351 & 0.227 &  0.118  &0.143  & 0.428 & 0.044 & 0.215 & 0.256 & 0.118 & 0.265 &0.543 \\
       \CMB &  \CMB     &         &     &0.162  &0.354 & 0.227  & 0.119 & 0.143 &0.428 & 0.044 & 0.216 & 0.257 &0.119 &0.266 &0.544 \\
       \CMB &  \CMB     &  \CMB   &     &0.167  & 0.358 &0.229 & 0.121 &0.147 &0.442 &0.045 &0.221 &0.263 &0.121 &0.272 &0.560 \\
       \CMB &   \CMB    &  \CMB  &  \CMB &0.173 & 0.368 & 0.240 &0.127 &0.157 &0.455 &0.047 &0.231 &0.276 &0.127 &0.290 &0.577 \\
       \toprule
    \end{tabular}
    }
    \caption{
    \footnotesize{
    Ablation studies on Apolloscape Car Instance test set.
     Superscript $S, M, L$ of average precision ($AP$s) and average recall ($AR$s) represent the object sizes. Superscript number $1, 10, 100$ represent the total number of detections for calculating recalls.}
    }\label{tab:ablation_apolloscape}
\end{table*}

\begin{table*}[t]
\vspace{-0.5cm}
\scriptsize{
    \centering%
    \begin{tabular}{@{}cccccc|cc@{}}
     \toprule
DHBM(w/o BWL) &|   DHBM  |  & GSI   |&NMR(w.\emph{R})|&  NMR| & GWME    & \multicolumn{2}{c}{\emph{``A3DP-Rel"} }    \\
             &             &        &                &       &          & private LB    &  public LB   \\ \hline
\CMB         &             &        &                &       &          & 0.108         &   0.115       \\
             & \CMB        &        &                &       &          & 0.112         &   0.118       \\
             & \CMB        & \CMB   &                &       &          & 0.122         &   0.128  \\
             & \CMB        & \CMB   & \CMB           &       &          & 0.127         &   0.129  \\
             & \CMB        & \CMB   &                & \CMB  &          & 0.130         &   0.136 \\
             & \CMB        & \CMB   &                &  \CMB & \CMB     & 0.142         &   0.150         \\

       \toprule
    \end{tabular}
 }
    \caption{
    \footnotesize{
    Ablation studies on Kaggle PKU/Baidu Dataset in terms of \emph{``A3DP-Rel"} criteria.
    DHBM(w/o BWL) denotes the dual head base model learning without the proposed Bayesian weight learning scheme.
    NMR(w.\emph{R}) denotes Neural 6-DoF Mesh Refiner including rotation matrix as optimisation variable.
    }
    }\label{tab:ablation_kaggle}
    \vspace{-1cm}
\end{table*}

\subsection{Review of the \emph{``A3DP-Rel"}  \&  \emph{``A3DP-Abs" } Metric:}
\begin{figure*}[t]
        \centering
        \includegraphics[width=1.0\textwidth]{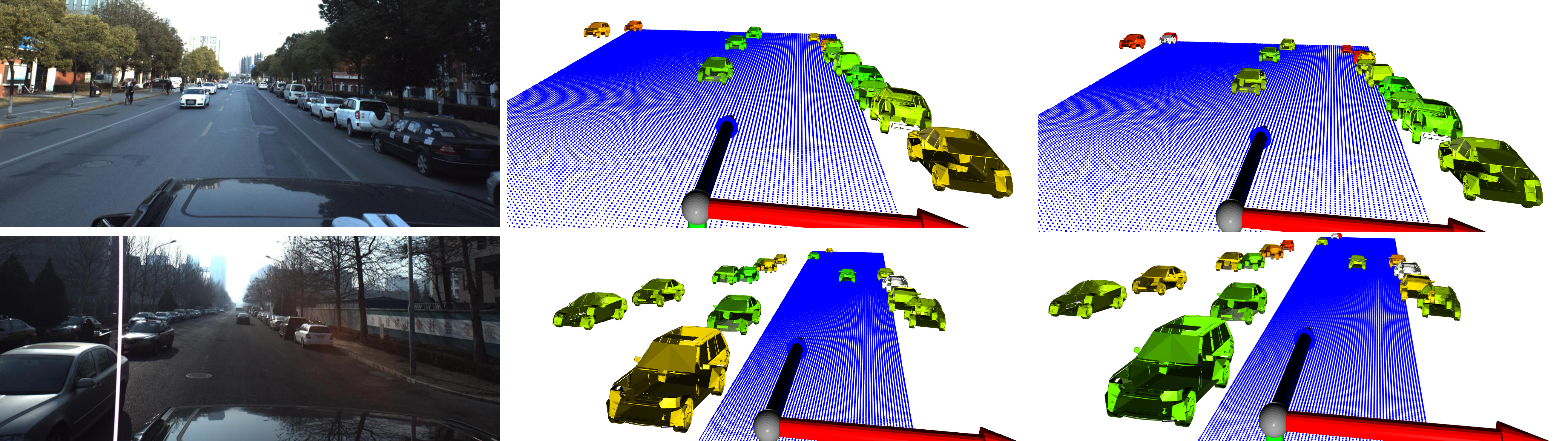}
        \caption{
        Differences for performance penalty for the \emph{``A3DP-Rel"} (middle) and the \emph{``A3DP-Abs" }(right) criteria.
        Left column: input monocular image, middle column and right column correspond to the same prediction overlayed with the aforementioned two metric criteria.
        Colour ``green"  denotes the true positive (the prediction satisfies the most strict $c_9$ criterion).
        Colour ``yellow" denotes the less accurate prediction mostly because of the translation estimation inaccuracy, and the more ``red" tone, the less accurate the prediction.
        Colour ``white" denotes the ground truth instance does not have any prediction correspond the most loose $c_0$ criteria (false negatives).
        }
        \vspace{-0.5cm}
        \label{fig:abs_res}
\end{figure*}
We conduct a critical review of the 3D metrics used to judge 3D monocular object detection for the two dataset.
In Fig.~\ref{fig:abs_res} we use Open3D~\cite{Zhou2018} for 3D visualising the same prediction with two criteria.
It is worth noting the strict translation distance threshold of 2.8 metres: it requires that the detected vehicle's distance from the camera centre needs to be correctly estimated within a 2.8 metres threshold even if the vehicle is hundreds metres away from the camera, otherwise the detection will be counted as a false positive for the \emph{``A3DP-Abs"} metric.
It can be seen that \emph{``A3DP-Rel"}  has a stricter criteria for the predictions close to the camera whereas  \emph{``A3DP-Abs"}  has a stricter criteria for predictions farther away from the camera. For the autonomous scenario, it's more advisable to pay more attention to the vehicles that are close to the camera. Hence, \emph{``A3DP-Rel"} is a more sensible metric in emphasising the contingency.

However, the \emph{``A3DP-Rel"} metric applies exactly eleven equally spaced recall levels, \emph{i.e.} $R_{11}=\{0, 0.1, 0.2, \ldots, 1\}$.
The interpolation function is defined as $\rho_{interp}(r)= \max_{r':r'\ge r \rho(r')}$, where $\rho(r)$ gives the precision at recall $r$, meaning that instead of averaging over
the actually observed precision values per point $r$, the maximum precision at recall value greater or equal than $r$ is taken. The recall intervals starts at 0, which means that a single , correctly matched prediction is sufficient to obtain 100\% precision at the bottom-most recall bin.
In the official implementation of \emph{``A3DP-Abs"}, the recall intervals start at 0, a sub-sampling 101 points is adopted. Hence, the miss detections are more penalised in  \emph{``A3DP-Abs"}. For example, changing the confident threshold from  $0.1$ to $0.9$  will lead to a increase of $3.8\%$ in \emph{``A3DP-Rel"} whereas an decrease of $7.0\%$ for \emph{``A3DP-Abs"}. In light of that false negatives could lead to disaster for autonomous driving, we should adopt the \emph{``A3DP-Abs"} scheme for calculating the recall.

\section{Conclusions}

We proposed a multi-staged 3D object detection deep neural network that can predict the instance's 3D shape and 6-DoF pose information.
By modelling homoscedastic uncertainty, we learn a dual-head base model to optimise the shape information, rotation and translation regression simultaneously.
Then we leverage the weakly supervised signal of 2D mask from the dual head prediction to refine the regressed 6-DoF pose estimation via a differentiable neural 3D mesh renderer.
To facilitate the mesh render to probe in a valid 3D space, we propose to initiate that state space by geometric reasoning.
We found our multi-stage scheme can help to achieve large improvement  over the direct regression neural architectures.
Our future direction is to explore the benefits of augmenting training images with synthetic data~\cite{alhaija2018augmented} to increase realism and diversity, especially for the rare cases that might never occur in the collected training set but are crucial for model generalisation.
Moreover, how to incorporate ambience lighting and instance's texture information into the optimisation process could be interesting future work.

\clearpage
%
%
\bibliographystyle{splncs04}
\bibliography{main}
\end{document}